\documentclass[journal]{IEEEtran}
\usepackage{amsmath,amsfonts}
\usepackage{algorithmic}
\usepackage{array}
\usepackage[caption=false,font=normalsize,labelfont=sf,textfont=sf]{subfig}
\usepackage{textcomp}
\usepackage{stfloats}
\usepackage{url}
\usepackage{verbatim}
\usepackage{graphicx}
\def\BibTeX{{\rm B\kern-.05em{\sc i\kern-.025em b}\kern-.08em
    T\kern-.1667em\lower.7ex\hbox{E}\kern-.125emX}}
\usepackage{balance}
\begin{document}
\title{SPDNet: Seasonal-Periodic Decomposition Network for Advanced Residential Demand Forecasting}

\author{Reza Nematirad, Anil Pahwa, Balasubramaniam Natarajan
}

\maketitle

\begin{abstract}
Residential electricity demand forecasting is critical for efficient energy management and grid stability. Accurate predictions enable utility companies to optimize planning and operations. However, real-world residential electricity demand data often exhibit intricate temporal variability, including multiple seasonalities, periodicities, and abrupt fluctuations, which pose significant challenges for forecasting models. Previous models that rely on statistical methods, recurrent, convolutional neural networks, and transformers often struggle to capture these intricate temporal dynamics. To address these challenges, we propose the Seasonal-Periodic Decomposition Network (SPDNet), a novel deep learning framework consisting of two main modules. The first is the Seasonal-Trend Decomposition Module (STDM), which decomposes the input data into trend, seasonal, and residual components. The second is the Periodical Decomposition Module (PDM), which employs the Fast Fourier Transform to identify the dominant periods. For each dominant period, 1D input data is reshaped into a 2D tensor, where rows represent periods and columns correspond to frequencies. The 2D representations are then processed through three submodules: a 1D convolution to capture sharp fluctuations, a transformer-based encoder to model global patterns, and a 2D convolution to capture interactions between periods. Extensive experiments conducted on real-world residential electricity load data demonstrate that SPDNet outperforms traditional and advanced models in both forecasting accuracy and computational efficiency. The code is available in this repository: https://github.com/Tims2D/SPDNet.
\end{abstract}


\section{Introduction}
\IEEEPARstart{E}{lectricity} demand forecasting is essential for the operation, control, and planning of modern power systems. Advancements in smart grid technologies and the widespread deployment of metering systems have enabled utility companies to access numerous high-resolution electricity consumption data at the residential level. This capability provides opportunities to forecast demand at individual household levels. Focusing on individual residential load forecasting offers significant advantages, such as more precise demand-side management, better monitoring of distribution systems, and enhanced integration of distributed energy resources and storage solutions at the community level \cite{10197224}.

Real-world residential electricity demand exhibits intricate temporal variations characterized by multiple seasonality, periodicity, and sharp fluctuations over short periods \cite{liu2024itransformer}. Multiple seasonality refers to the presence of multiple recurring patterns within the data that occur at different regular intervals, such as daily, weekly, and monthly. Periodicity pertains to recurring patterns or cycles that repeat over fixed intervals \cite{anonymous2024timesd}. The electricity demand is influenced by both within-period and between-period correlations. Within-period correlations occur when the electricity demand at one time is influenced by the demand at another time within the same period. Between-period correlations involve associations of electricity demand at the same time across different periods. Additionally, sharp fluctuations over short periods are common in residential electricity demand. These abrupt changes can result from sudden variations in household activities, such as turning on high-power appliances, or from external factors like weather-induced heating or cooling demands \cite{nematirad2024optimization}. These complexities pose significant challenges for residential load forecasting. To address these challenges, utility companies typically aggregate electricity demand from consumers connected to a single substation or transformer. This aggregation smoothes out fluctuations in residential electricity load, facilitating more accurate forecasting \cite{lei2024privgrid}. However, recent advances in data acquisition technology at the household level and the shift toward smart communities have increased the need for forecasting individual household electricity demand. Aggregated load demand forecasting has been extensively investigated, resulting in acceptable levels of accuracy \cite{bashiri2024review}. Conversely, residential demand forecasting remains challenging due to higher variability and less predictable consumption patterns. Consequently, advanced methods have been developed for individual residential electricity demand in recent years.

Traditional statistical models, such as Autoregressive Integrated Moving Average (ARIMA) and Exponential Smoothing methods, have been widely used for load forecasting. ARIMA models rely on the assumption that future values are linear functions of past observations and past errors \cite{tarmanini2023short}. Exponential Smoothing methods are particularly suitable for time series data exhibiting trends and seasonality \cite{dudek2021hybrid}. Although ARIMA and Exponential Smoothing models perform well with linear and seasonal electricity data, they struggle with complex, nonlinear, and non-stationary time series data \cite{wu2023timesnet}. Multilayer Perceptrons (MLPs) have been developed to overcome these limitations. MLPs consist of multiple layers of interconnected neurons with nonlinear activation functions, allowing them to learn complex mappings from inputs to outputs \cite{nematirad2022solar}. However, MLPs lack the inherent ability to model temporal dependencies in sequential data because they do not have mechanisms to handle sequences or memory of previous inputs \cite{faustine2024efficiency}. 

For modeling temporal sequence dependency, advanced deep learning models have been developed. These models by leveraging hierarchical feature learning, non-linear modeling, high-dimensional pattern recognition, and adaptive learning capabilities, can model the intricate patterns present in residential electricity demand \cite{chen2023short}. For instance, Recurrent Neural Networks (RNNs) and their variants are designed to process sequential data by maintaining a hidden state that captures information from previous time steps. This enables them to model temporal dependencies effectively. However, RNNs often struggle with long-term dependencies due to vanishing or exploding gradients. Long Short-Term Memory (LSTM) networks, a type of RNN, are designed to address the vanishing or exploding gradient problem by incorporating gating mechanisms that regulate the flow of information through the memory cells. However, LSTMs may not efficiently capture local features and patterns that occur within short periods \cite{kag2020rnns}. 

Convolutional Neural Networks (CNNs) have been adapted to capture local patterns through convolutional filters that slide over input sequences \cite{nematirad2024acoustic}. These filters enable CNNs to detect and learn dependencies between adjacent time points. However, standard CNNs have limitations in capturing long-term dependencies due to their fixed receptive fields, which are determined by kernel size and network depth \cite{shi2023load}. To address this limitation, architectures like Convolutional LSTMs (ConvLSTMs) combine the strengths of CNNs and LSTMs by integrating convolutional operations within the LSTM gates \cite{xiao2024improved}. Alternatively, Temporal Convolutional Networks (TCNs) employ causal and dilated convolutions to capture long-range temporal dependencies while efficiently modeling local patterns. Despite this advantage, TCNs may struggle to model complex nonlinear interactions between distant time points and irregular temporal patterns \cite{xu2023phacia}. 

Recently, transformer models have been developed to effectively capture both local and global dependencies in sequential data \cite{zhou2021informer}. Transformers utilize a self-attention mechanism, enabling the model to weigh the importance of different time steps when making predictions. Unlike recurrent models, transformers do not rely on the sequential processing of data, allowing for greater parallelization and efficiency during training. Despite these advantages, transformers can be computationally intensive and often require large amounts of data to achieve optimal performance \cite{fan2024optimizing}.

\begin{figure*}[t]
\centering
\includegraphics[width=\textwidth]{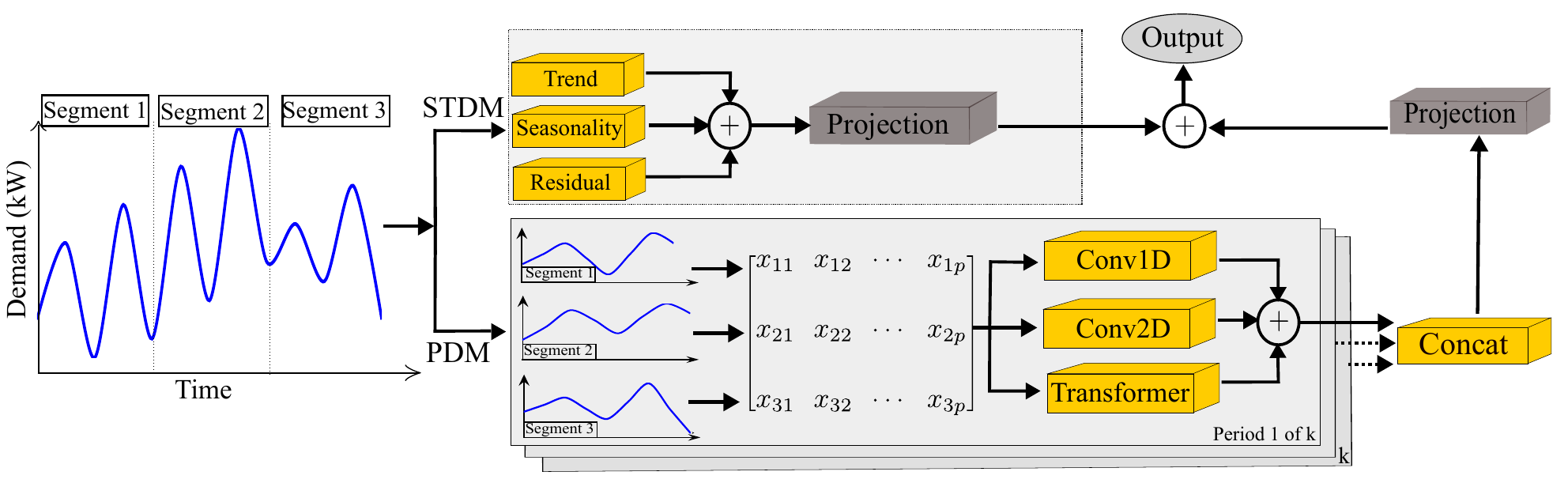} 
\caption{Overview of the proposed Seasonal-Periodic Decomposition Network (SPDNet) for residential electricity demand forecasting.}
\label{Figure1}
\end{figure*}

Motivated by the limitations of existing forecasting models, this paper introduces SPDNet, a novel framework for residential electricity load forecasting. SPDNet is a modification of Times2D \cite{anonymous2024timesd}, our previous model for general time series forecasting. The modification addresses the seasonality and trend with a distinct process, and local dependencies utilizing convolution layers, whereas in Times2D first and second derivatives are used. 
SPDNet consists of two main modules: the Seasonal-Trend Decomposition Module (STDM) and the Periodical Decomposition Module (PDM). The STDM decomposes the input data into trend, seasonal, and residual components to capture underlying temporal patterns effectively. The PDM transforms 1D inputs into 2D representations through frequency domain analysis. Utilizing the Fast Fourier Transform (FFT), the PDM identifies \( k \) dominant period-frequency pairs within the input data. Each pair is then used to reshape the 1D series into a corresponding 2D tensor. The rows of each tensor represent the period, while the columns align time points across this period. As a result, the input 1D time series is transformed into \( k \) distinct 2D tensors, each corresponding to one of the identified dominant periods. The PDM includes three submodules that are applied in parallel to each 2D tensor iteratively: (1) a short-term modeling submodule empowered by 1D convolutions to capture sharp fluctuations within short periods, (2) a transformer-based encoder to model global patterns within periods, and (3) a 2D convolutional submodule to discover correlations between and within periods. Figure~\ref{Figure1} presents the systematic workflow of the proposed model.
The contributions of this paper can be summarized as follows:
\begin{itemize}
    \item \textbf{Proposed a novel framework}: Introducing SPDNet, a deep learning framework designed to model complex temporal patterns for residential electricity load forecasting effectively.
    \item \textbf{Seasonal-Trend Decomposition Module (STDM)}: Developing the STDM to decompose load data into trend, seasonal, and residual components, enhancing the capability to model long-term trends and repeating patterns.
\item \textbf{Integrated specialized submodules}: PDM incorporates three specialized submodules to model various temporal patterns effectively. The 1D Convolutional submodule captures sharp fluctuations within short periods, modeling abrupt changes in consumption behavior. The Transformer-based Encoder submodule models global patterns within periods, capturing long-range dependencies and complex temporal relationships, which are crucial for understanding broader trends. Finally, the 2D Convolutional submodule uncovers relationships both between and within periods, enabling the model to capture inter-period and intra-period correlations, enhancing its ability to represent periodic dependencies comprehensively.
    \item \textbf{Achieved state-of-the-art performance}: SPDNet outperforms existing models on real-world data, achieving comprehensive state-of-the-art results in both forecasting accuracy and computational efficiency.
\end{itemize}

\subsection{Related works}
\subsubsection{Recurrent Neural Networks}
RNNs and their variants have been widely employed for electricity load forecasting. For example, RNNs are utilized in \cite{morais2023short} for electricity load forecasting for the New York Independent System Operator. The study compared Deep-RNN with a Deep Feedforward Neural Network and showed that Deep-RNN achieved better performance for both 24-hour and 168-hour horizons. A Gated Recurrent Unit (GRU) model is used in \cite{aseeri2023effective} for day-ahead multivariate demand forecasting, focusing on large-scale power grid data. The results showed that the GRU-based approach significantly outperformed statistical methods, such as ARIMA. An LSTM model for univariate load prediction is utilized in \cite{bouktif2018optimal}. The results demonstrated that LSTM outperformed traditional approaches such as Support Vector Machines (SVM) and decision trees, achieving better performance in short-term forecasts. A bidirectional Long Short-Term Memory (BiLSTM) model is employed in \cite{zhang2024forecasting} for forecasting building electricity consumption. Based on the simulation, the BiLSTM improved prediction accuracy by 23.97\% compared to LSTM method. \cite{10679177} proposed an improved 3D LSTM model for load forecasting by converting time series data into 3D continuous video frames. 3D-LSTM demonstrated superior performance compared to advanced models like Transformers. In addition, a novel short-term load forecasting framework based on a parallel CNN-LSTM network is proposed in \cite{kshetrimayum2024pconvlstm} to capture temporal and spatial patterns. Despite their success, RNNs and their variants suffer from vanishing gradients and capturing local patterns.

\subsubsection{Multi-Layer Perceptron}
An MLP-based model is proposed in \cite{10529636} for net-load forecasting in low-voltage distribution networks. Temperature and solar irradiance are used as exogenous variables. The proposed MLP model outperforms models like LSTM and random forests (RF) in computational efficiency and forecast accuracy. A Feed-Forward MLP is used in \cite{10649928} for sub-hourly load forecasting. The study considered electric vehicles and AC loads as exogenous parameters. Results showed that the proposed model outperformed the XGBoost and Prophet models. Neural Basis Expansion Analysis Time Series (N-BEATS) is proposed in \cite{singh2024short} that primarily uses MLPs for short-term forecasting in smart grids. Comparative results demonstrated that N-BEATS outperforms ARIMA, SVM, RF, CNN, and BiLSTM. Neural Hierarchical Interpolation for Time Series (N-HiTS) and Transformer are utilized in \cite{liang2024assessing} for power demand forecasting. In forecasting performance, the Transformer and N-HiTS performed similarly, but N-HiTS was more computationally efficient in handling longer input data. Despite their advantages, MLPs lack the inherent ability to model temporal dependencies in sequential data because they do not have mechanisms to handle sequences or retain the memory of previous inputs.

\subsubsection{Convolutional Neural Networks}
A CNN model is introduced in \cite{ghimire2023novel} for daily electricity demand prediction. Based on the results, the model outperformed benchmark models like SVR, MLP, and RNN across multiple substations in Southeast Queensland, Australia. A CNN-LSTM hybrid model is proposed in \cite{10111057} for electricity load forecasting. The CNN captures local temporal patterns, while the LSTM decoder models long-term dependencies. A CNN-BiLSTM model has been proposed in \cite{10530268} for residential electricity demand forecasting. The results highlight the advantage of combining CNN with BiLSTM, particularly in high-frequency residential electricity demand data. TCNs are introduced in \cite{turkouglu2024integrated} to improve electricity demand forecasting in residential settings. The study demonstrated that the TCN model outperforms models such as LSTM and GRU. 
A hybrid model that leverages CNN, transformer, and LSTM is proposed in \cite{he2024enhancing} for short-term power load forecasting. In this model, 2D CNN layers convert one-dimensional time series data into two-dimensional space for better feature extraction. The proposed model showed superior performance compared to traditional models such as CNN, TCN, and LSTM. Despite their effectiveness in capturing local temporal patterns, CNNs have limitations in capturing long-term dependencies. 

\subsubsection{Transformers}

A transformer-based approach to forecasting individual household load demand by employing a sparse attention mechanism to address the limitations of RNNs and CNNs is proposed in \cite{chan2024transformer}. The model leverages historical electricity load and weather data as exogenous variables. The results demonstrate superior computational efficiency and accuracy compared to RNN and LSTM models. The Informer network \cite{zhou2021informer} combined with time2vec embedding and 1D convolutional layers is proposed in \cite{huang2024day} for day-ahead load forecasting. Temperature and dew point are used as exogenous inputs to capture periodic and non-periodic patterns, which makes this model more accurate and efficient than LSTMs and Transformers. \cite{li2024short} proposed a Crossformer model \cite{zhang2023crossformer} that integrates a hierarchical encoder-decoder structure and variational mode decomposition to improve short-term electricity load forecasting. The model focuses on cross-dimensional dependencies in multivariate data to improve prediction accuracy compared to other transformer models. PatchTST, an advanced transformer-based model integrated with TCN, is used for multi-load forecasting in smart buildings in \cite{10403885}. The results demonstrate the superior performance and computational efficiency of PatchTST over models such as LSTM, GRU, RNN, Transformer, and Informer. Despite their advantages, transformers are computationally demanding and typically require substantial amounts of data to achieve optimal performance.

\section{SPDNet Methodology}
The following subsections outline the steps involved in the proposed SPDNet for electricity demand forecasting.
\subsection{Data Preparation and Reshaping}
Electricity demand data is often used in conjunction with auxiliary variables to enhance forecast accuracy. This dataset is represented as \( X = [x_1, x_2, \dots, x_T] \in \mathbb{R}^{T \times N} \), where \( T \) denotes the length of the time series and \( N \) represents the number of variables. For effective analysis using deep learning models, the dataset must be properly formatted by defining the input and output sequences. First, a segment of the time series, referred to as the sequence length \( S \), is selected as the input. A subsequent segment of the time series, referred to as the prediction length \( P \), is selected as the output. By following this approach, the original data of total length \( T \) is transformed into multiple rows, where each row contains \( S \) input time points and \( P \) corresponding output. Additionally, the data is organized into batches for parallel processing during training. The samples are organized into batches of size \( B \), resulting in input tensors with dimensions \( [B, S, N] \) and output tensors with dimensions \( [B, P, N] \). 

\subsection{Seasonal-Trend Decomposition Module}
STDM is designed to extract the trend, seasonality, and residual components from electricity demand data. Given an input tensor \( X \in \mathbb{R}^{B \times S \times N} \), the decomposition is performed as follows:
\begin{equation}
X_{\text{trend}}, X_{\text{seasonal}} = \text{STDM}(X)
\end{equation}
The trend component \( X_{\text{trend}} \) is extracted using a one-dimensional convolution that captures the long-term trends in the time series. The trend is computed as
\begin{equation}
X_{\text{trend}} = \text{Conv1D}_{\text{trend}}(X)
\end{equation}
Here, the \( \text{Conv1D}_{\text{trend}} \) employs a large kernel size, which is suitable for capturing long-term trends. The convolution is performed along the temporal dimension \( S \). The seasonality component captures the repeating patterns that occurs at specific intervals. It is extracted by applying a convolution to the detrended data, which is the original data minus the trend component, as follows:
\begin{equation}
X_{\text{seasonal}} = \text{Conv1D}_{\text{seasonal}}(X - X_{\text{trend}})
\end{equation}
In this case, the convolution operation \( \text{Conv1D}_{\text{seasonal}} \) employs an appropriate kernel size, typically smaller than the trend kernel. The residual component \( X_{\text{residual}} \) represents the irregularities or noise that are not explained by the trend or seasonality components. It is computed by subtracting the extracted trend and seasonality components from the original input:
\begin{equation}
X_{\text{residual}} = X - X_{\text{trend}} - X_{\text{seasonal}}
\end{equation}
Finally, the output of the STDM, denoted as \( X_{\text{STDM}} \), is computed by summing the seasonality, trend, and residual components:
\begin{equation}
{X}_{\text{STDM}} = X_{\text{seasonal}} + X_{\text{trend}} + X_{\text{residual}}
\end{equation}
To prepare the enhanced temporal features \( X_{\text{STDM}} \in \mathbb{R}^{B \times S \times N} \) extracted from the decomposition for the prediction, the model employs a linear projection layer that transforms the sequence length \( S \) of the input data into the prediction length \( P \) required for forecasting as
\begin{equation}
\hat{X}_{\text{STDM}} = \text{LinearProjection}(X_{\text{STDM}})
\end{equation}
where \( \hat{X}_{\text{STDM}} \in \mathbb{R}^{B \times P \times N} \) denotes the processed representation obtained from the STDM, which serves as the input to subsequent layers for the final forecasting. 

\subsection {Periodical Decomposition Module}
Parallel with the STDM, the PDM is designed to extract periodic patterns from the data. The objective is to break down the 1D input tensor \( X \in \mathbb{R}^{B \times S \times N} \) into multiple 2D periodic components to capture short-term variations, inter and intra-period patterns, and long-term behaviors across different periods. To achieve this, a FFT is applied along the temporal dimension \( S \) of the input tensor \( X \in \mathbb{R}^{B \times S \times N} \) for each batch and feature to identify the dominant periods as follows:
\begin{equation}
X_f =  \sum_{t=0}^{S-1} X(t) e^{-2 \pi i t f / S}
\end{equation}
Here, \( X_f \) represents the transformed tensor in the frequency domain and \( f \) is the frequency. The magnitude \( A_f \) can be calculated as
\begin{equation}
A_f = |X_f| = \sqrt{\text{Re}(X_f)^2 + \text{Im}(X_f)^2}
\end{equation}
Since \( X_f \) is symmetric, we maintain only the first half of the frequency components, corresponding to the non-negative frequencies. Accordingly, the magnitude \( A_f \) becomes a tensor of dimensions \( [B, \frac{S}{2}, N] \).

\begin{figure}[t]
\centering
\includegraphics[width=\columnwidth]{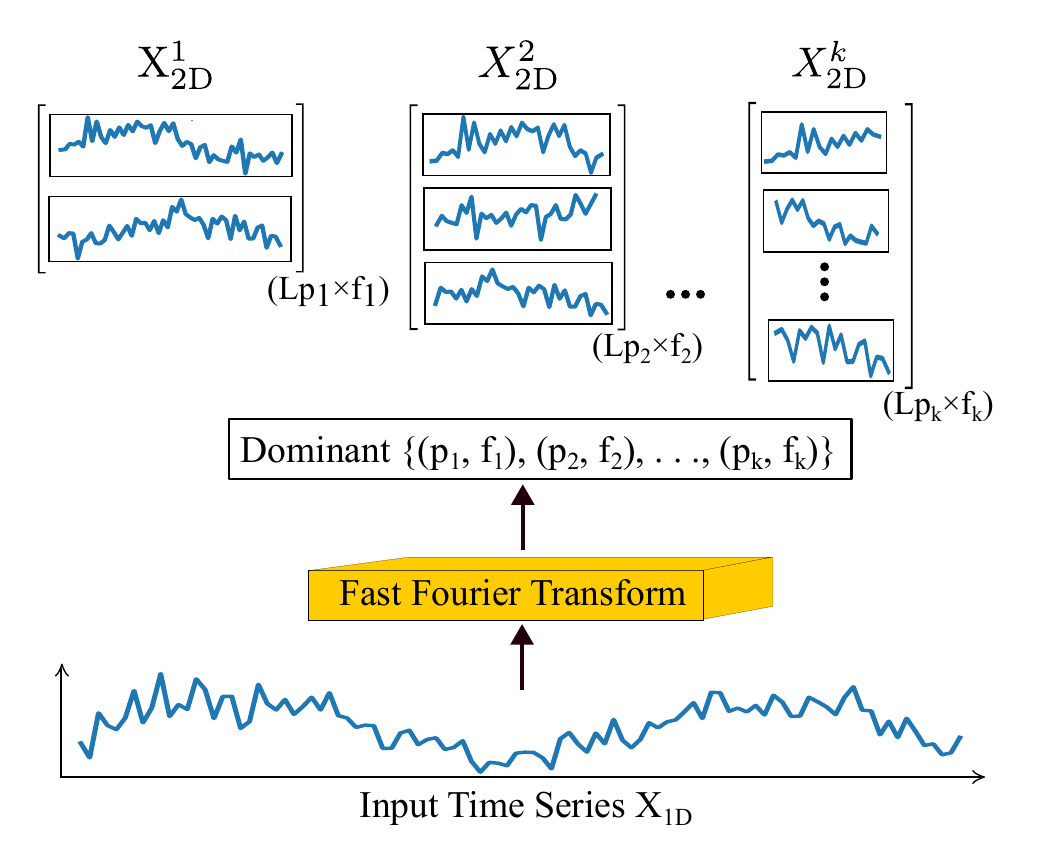} 
\caption{Overview of the Periodical Decomposition Module (PDM) Framework to reshape the 1D input tensor into \( k \) distinct 2D tensors.}
\label{Figure2}
\end{figure}

To identify the dominant periods for each sequence, the magnitudes are averaged over the \( B \) and \( N \) dimensions. The resulting magnitudes are then sorted in descending order, and the dominant frequencies are selected based on their associated magnitudes. It should be noted that the highest magnitude corresponds to the zero frequency, representing the DC component, which indicates the overall mean and is not considered for periodic analysis. As a result, the first \( k \) dominant periods \( p_1, p_2, \dots, p_k \), associated with the dominant frequencies \( f_1, f_2, \dots, f_k \), can be calculated as
\begin{equation}
P_k = \frac{S}{f_k}
\end{equation}
Thus, the original 1D input tensor \( X \in \mathbb{R}^{B \times S \times N} \) can be decomposed into \( k \) distinct 2D tensors \( X_{\text{2D}}^i \in \mathbb{R}^{B \times p_i \times f_i \times N} \), each corresponding to one of the identified dominant periods. For instance, the \( i \)-th 2D tensor has the dimensions \( [B, p_i, f_i, N] \), where \( p_i \) represents the number of full periods within the sequence, and \( f_i \) denotes the length of each period. It should be noted that to preserve the structural integrity of the data during decomposition, areas where the product of period length \( f_i \) and the number of periods \( p_i \) does not fully cover the sequence length \( S \) are padded with zeros. This padding ensures that the input tensor \( X \in \mathbb{R}^{B \times S \times N} \) can be evenly divided into \( k \) distinct 2D tensors \( X_{\text{2D}}^i \in \mathbb{R}^{B \times p_i \times f_i \times N} \). Figure~\ref{Figure2} illustrates the transformation process of the original 1D input tensor into multiple 2D tensors. Then, the 2D periodic tensors \( X_{\text{2D}}^i \) are used to capture short-term variations, inter- and intra-period dependencies, and long-term trends.

\subsubsection{Short-term Analysis} For short-term cyclic behavior analysis, the decomposed 2D tensors \( X_{\text{2D}}^i \) are processed to capture sharp fluctuations within the identified periods efficiently as the following step:\\
\textbf{Reshaping}: Given an input tensor \( X_{\text{2D}}^i \in \mathbb{R}^{B \times p_i \times f_i \times N} \), it is reshaped by merging the \( B \) and \( N \) dimensions, resulting in \( X_{\text{reshaped}}^i \in \mathbb{R}^{(B \times N) \times p_i \times f_i} \) as

\begin{equation}
X_{\text{reshaped}}^i = \text{Reshape}(X_{\text{2D}}^i, [B \times N, p_i, f_i])
\end{equation}
This reshaping is necessary to facilitate efficient convolutional operations over the time dimension.\\
\textbf{1D Convolution}: A one-dimensional convolution is then applied over the temporal dimension of the reshaped tensor \( X_{\text{reshaped}}^i \). This operation is designed to capture short-term patterns or fluctuations within each identified period as
\begin{equation}
X_{\text{short}}^i = \text{Conv1D}(X_{\text{reshaped}}^i)
\end{equation}
\textbf{Projection}: After extracting short-term patterns, the output tensor is projected to the desired prediction length \(P\). A linear projection layer is employed to prepare the feature representations for forecasting as
\begin{equation}
\hat{X}_{\text{short}}^i = \text{LinearProjection}(X_{\text{short}}^i)
\end{equation}
where \(\hat{X}_{\text{short}}^i\) with shape \([B, P, N]\), denotes the short or local extracted features, which serve as the input to subsequent layers.
\subsubsection{Periodic Dependency Analysis}
In parallel with the short-term analysis, the periodic dependency analysis module processes the 2D tensors \( X_{\text{2D}}^i \) to capture both intra-period and inter-period dependencies simultaneously. The following steps outline the procedure:\\
\textbf{Reshaping}: The periodic tensors \( X_{\text{2D}}^i \in \mathbb{R}^{B \times p_i \times f_i \times N} \) undergo a two-step reshaping process. Initially, similar to Equation (10) for short-term analysis, the \( B \) and \( N \) dimensions are merged, forming \( X_{\text{reshaped}}^i \). Subsequently, an extra dimension is introduced to facilitate two-dimensional analysis as
\begin{equation}
X_{\text{reshaped}}^i = \text{unsqueeze}(X_{\text{reshaped}}^i)
\end{equation}
This process reshapes the tensor into \(X_{\text{2D}}^i\) with dimensions \([B \times N, 1, p_i, f_i]\). By adding this extra dimension, the 2D convolution is able to simultaneously capture intra-period dependencies across the \(p_i\) dimension and inter-period dependencies across the \(f_i\) dimension.\\
\textbf{2D Convolution}: A two-dimensional convolution is applied to the reshaped tensor \(X_{\text{reshaped}}^i \in \mathbb{R}^{(B \times N)\times 1\times p_i \times f_i}\) as
\begin{equation}
\hat{X}_{\text{2D}}^i = \text{Conv2D}(X_{\text{reshaped}}^i)
\end{equation}
\textbf{Projection}:
After 2D convolution, the tensor \(\hat{X}_{\text{2D}}^i\) is processed through a projection layer to map the features to the desired prediction length \( P \), i.e.,
\begin{equation}
\hat{X}_{\text{Periodic}}^i = \text{LinearProjection}(\hat{X}_{\text{2D}}^i)
\end{equation}
where \(\hat{X}_{\text{Periodic}}^i\) with shape \([B, P, N]\), denotes the extracted features representing periodic dependencies, which serve as the input to subsequent layers of the model. 
\subsubsection{Long-term Analysis}
This module is designed to model global or long-term dependencies that influence future values. The following steps outline the procedure:\\
\textbf{Reshaping}: Similar to Equation (10) for short-term analysis, the \( B \) and \( N \) dimensions of the input 2D tensors \( X_{\text{2D}}^i \in \mathbb{R}^{B \times p_i \times f_i \times N} \) are merged, resulting in the reshaped tensor \( X^i \in \mathbb{R}^{(B \times N) \times p_i \times f_i} \). This tensor is then used in subsequent processing steps.\\
\textbf{1D Convolution}: A Conv1D is applied over the temporal dimension of the reshaped tensor \( X^i \in \mathbb{R}^{(B \times N) \times p_i \times f_i } \). This step is crucial for extracting and enriching temporal patterns within each period as follow: 
\begin{equation}
X_{\text{conv}}^i = \text{Conv1D}(X^i)
\end{equation}
\textbf{Reshaping}: After enriching input data with Conv1D, the \(X_{\text{conv}}^i\) is reshaped back to shape \([B, S, N]\) as:
\begin{equation}
X_{\text{reshaped}}^i = \text{reshaped}(X_{\text{conv}}^i, [B, S, N])
\end{equation}
This step realigns the tensor for embedding processes.\\
\textbf{Reverted Embedding}: This unique embedding layer processes each time series independently, operating on the rows of the reshaped tensor \( X_{\text{reshaped}}^i \in \mathbb{R}^{B \times S \times N} \). Unlike traditional embedding methods that map along the feature dimension \( N \), the reverted embedding maps along the temporal sequence \( S \). This embedding utilizes a series of linear layers to project each time series into a higher-dimensional \( d_{\text{model}} \) as \cite{liu2024itransformer}
\begin{equation}
X_{\text{embedded}}^i = \text{RevertedEmbedding}(X_{\text{reshaped}}^i)
\end{equation}
Here, \(X_{\text{embedded}}^i\) with shape \([B, d_{\text{model}}, N]\) represents the higher-dimensional feature representation of each time series. This embedding facilitates a deeper analysis by enhancing the ability of the model to discern long-term temporal dependencies, preparing the data for sophisticated processing with transformer encoder layers.\\
\textbf{Encoder Module}: The encoder processes the embedded tensors sequentially through \(L\) stacked layers, each containing key submodules such as multi-head attention mechanisms, Feed-Forward Networks (FFN), residual connections, and normalization layers. Each component plays a crucial role in enhancing the data processing capabilities of the model. At each encoder layer \( \ell \), the input \(X_{\text{embedded}}^i\) is transformed through a series of operations within the encoder. Initially, the Queries (\(Q_{\ell}^{(h,i)}\)), Keys (\(K_{\ell}^{(h,i)}\)), and Values (\(V_{\ell}^{(h,i)}\)) for each input \(i\) are computed at layer \( \ell \) as follows:
\begin{align}
Q_{\ell}^{(h,i)} &= W_{Q_{\ell}}^{(h,i)} X_{\text{embedded}}^i \\
K_{\ell}^{(h,i)} &= W_{K_{\ell}}^{(h,i)} X_{\text{embedded}}^i \\
V_{\ell}^{(h,i)} &= W_{V_{\ell}}^{(h,i)} X_{\text{embedded}}^i
\end{align}
Here, \( W_{Q_{\ell}}^{(h,i)} \), \( W_{K_{\ell}}^{(h,i)} \), and \( W_{V_{\ell}}^{(h,i)} \) are learnable weight matrices specific to each head \( h \), each layer \( \ell \), and each input tensor \( i \). Here, the queries (\(Q_{\ell}^{(h,i)}\)) indicate what each time step aims to learn from the others, the keys Keys (\(K_{\ell}^{(h,i)}\)) provides information about the time steps, and the values (\(V_{\ell}^{(h,i)}\)) contain the corresponding data, which is weighted based on the attention scores derived from comparing queries and keys.

Next, the multi-head self-attention mechanism is employed to determine the relative importance of each time step within the sequence. For each head \(h\), attention weights are computed by taking the dot product of the Queries \(Q_{\ell}^{(h,i)}\) and Keys \(K_{\ell}^{(h,i)}\), scaling the result by the inverse square root of the dimension of the keys \(d_k\), and then applying a softmax function to normalize the scores. This operation can be mathematically expressed as \cite{liu2024itransformer}
\begin{equation}
\text{A}_{\ell}^{(h,i)} = \text{softmax}\left(\frac{Q_{\ell}^{(h,i)} K_{\ell}^{(h,i)T}}{\sqrt{d_k}}\right) V_{\ell}^{(h,i)}
\end{equation}
where \(\text{A}_{\ell}^{(h,i)}\) represents the attention matrix for head \(h\) regarding the \(i\)-th input at layer \(\ell\). This operation calculates how much focus each time step should place on others, allowing the model to identify relationships between different points in the sequence. Each head operates independently using a different set of learnable weights, enabling the model to capture a variety of relationships within the sequence. 
Then the outputs of the \(h\) attention mechanisms are concatenated over the feature dimension as
\begin{equation}
X_{\text{Concat},\ell}^{i} = \text{Concat}(A_{\ell}^{(1,i)}, A_{\ell}^{(2,i)}, \dots, A_{\ell}^{(H,i)})
\end{equation}
where \(X_{\text{Concat},\ell}^{i}\) denotes the concatenated output of the multi-head attention mechanism for the \(i\)-th input at layer \(\ell\).
The concatenated output is then passed through a Feed-Forward Network (FFN) to introduce non-linearity, enhancing the ability of the model to capture complex patterns, i.e.,
\begin{equation} 
X_{\text{FFN},\ell}^{i} = \text{FFN}(X_{\text{Concat},\ell}^{i})
\end{equation}
where \(X_{\text{FFN},\ell}^{i}\) represents the output from the FFN at layer \(\ell\) for the \(i\)-th input. To maintain the integrity of the original input data and enhance feature integration, a residual connection is added to the FFN output. This step helps preserve information from earlier layers and facilitates deeper network training without the risk of vanishing gradients. The residual connection sums the input to the layer and the output from the FFN is
\begin{equation}
X_{\text{residual},\ell}^{i} = X_{\text{Concat},\ell}^{i} + X_{\text{FFN},\ell}^{i}
\end{equation}
where \(X_{\text{residual},\ell}^{i}\) represents the residual output at layer \(\ell\) for the \(i\)-th input. This residual output is then normalized to stabilize the transformed output:
\begin{equation}
X_{\text{normalized},\ell}^{i} = \text{Normalization}(X_{\text{residual},\ell}^{i})
\end{equation}
where \(X_{\text{normalized},\ell}^{i}\) denotes the output of the normalization process at layer \(\ell\) for the \(i\)-th input.

\begin{figure*}[t]
\centering
\includegraphics[width=\textwidth]{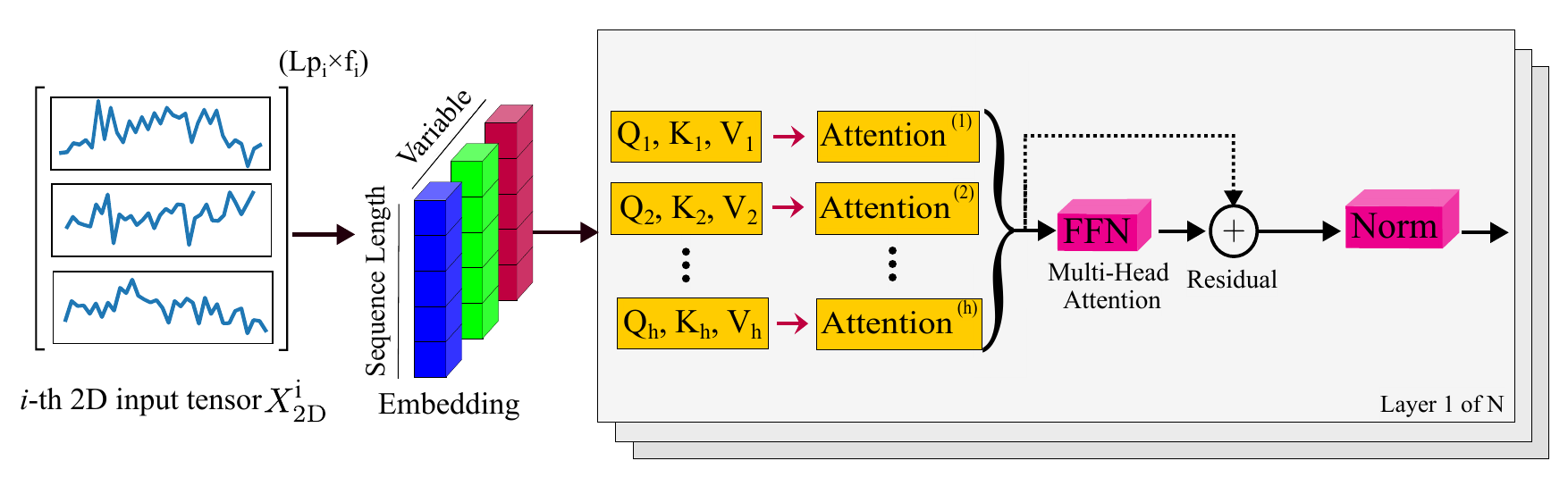} 
\caption{Overall architecture of transformer encoder processing to capture global dependencies.}
\label{Figure3}
\end{figure*}

This encoder process consists of multi-head self-attention mechanisms, connections, FFN, residual connections, and normalization are repeated over \(L\) layers in the encoder. After all \(L\) layers have been processed, the final encoded representation \(X_{\text{encoded}}^i\) for the \(i\)-th periodic input tensor \(X^i\) is obtained.
At the end, \(X_{\text{encoded}}^i\) is processed through a projection layer to map the features to the desired prediction length \(P\):

\begin{equation}
\hat{X}_{\text{long}}^i = \text{LinearProjection}(X_{\text{encoded}}^i)
\end{equation}
Here, \(\hat{X}_{\text{long}}^i\) with shape \([B, P, N]\) denotes the extracted long-term or global dependencies, which serve as the input to subsequent layers of the model. In addition, the overall architecture of the encoder processing is depicted in Figure~\ref{Figure3}. 

\subsubsection{Summation}
The PDM divides the 1D input tensor \(X \in \mathbb{R}^{B \times S \times N}\) into \(k\) distinct 2D tensors \(X^i_{\text{2D}} \in \mathbb{R}^{B \times p_i \times f_i \times N}\). Each tensor \(X^i_{\text{2D}}\) is processed through short-term analysis, periodic dependency analysis, and long-term analysis to capture the relevant features from the \( i \)-th dominant period independently.The outputs from these analyses, \(\hat{X}_{\text{short}}^i\), \(\hat{X}_{\text{Periodic}}^i\), and \(\hat{X}_{\text{long}}^i\) should be combined to create a comprehensive representation of the different temporal scales inherent in the data. This integration is performed through a summation as

\begin{equation} 
\hat{X}_{\text{PDM}}^i = \hat{X}_{\text{short}}^i + \hat{X}_{\text{periodic}}^i + \hat{X}_{\text{long}}^i 
\end{equation}
After summation, the integrated outputs \(\hat{X}_{\text{PDM}}^i\) with shape \([B,P,N]\) should be fused for \(k\) different dominant periods as

\begin{equation} 
\hat{X}_{\text{PDM\_concat}} = \text{Concat}(\hat{X}_{\text{PDM}}^1, \ldots, \hat{X}_{\text{PDM}}^k) 
\end{equation}
where \(\hat{X}_{\text{PDM\_concat}}\) with shape \([B, P, N, K]\) represents the comprehensive temporal features aggregated across all \(k\) dominant periods.
Finally, to refine and consolidate these features for prediction, a linear transformation is applied to collapse the additional dimension \(k\) introduced by concatenation, i.e.,
\begin{equation}
\hat{X}_{\text{PDM}} = \text{LinearTransform}(\hat{X}_{\text{PDM\_concat}})
\end{equation}
where \(\hat{X}_{\text{PDM}}\) with shape \([B, P, N]\) denotes the final processed output of the PDM.

\subsection{Aggregation}
The final output of the PDM \( \hat{X}_{\text{PDM}} \) and the output from the STDM \( \hat{X}_{\text{STDM}} \) are then combined through a weighted addition to create a final feature representation for the forecasting task. This process is expressed mathematically as follows:

\begin{equation} 
\hat{X} = \alpha_1 \hat{X}_{\text{PDM}} + \alpha_2 \hat{X}_{\text{STDM}}
\end{equation}
where \( \alpha_1 \) and \( \alpha_2 \) are learnable weights that adjust the influence of periodical dependencies and seasonal patterns, respectively. The combined output \(\hat{X}\) with shape \([B, P, N]\) encapsulates various temporal variations captured by the PDM and STDM. This representation is subsequently passed through a loss function during the training phase to optimize the model parameters for accurate forecasting.

\begin{figure*}[t]
\centering
\includegraphics[width=\linewidth]{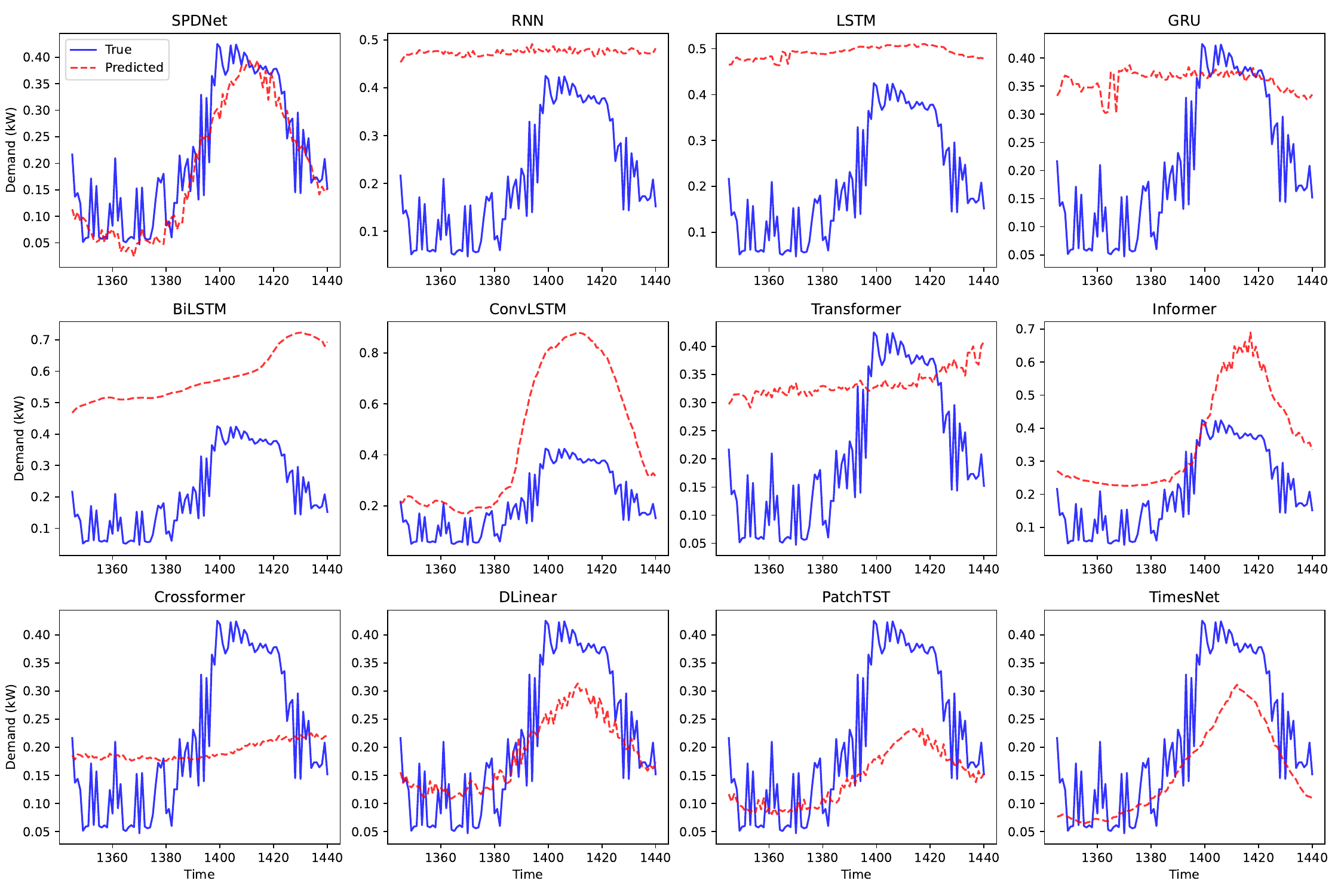} 
\caption{Forecasted and actual electricity demand for Load 1 at sequence length $S=96$ and prediction horizon $P=96$. Legends are shown only in the first subplot for clarity.}

\label{Figure4}
\end{figure*}

\begin{figure}[t]
\centering
\includegraphics[width=\linewidth]{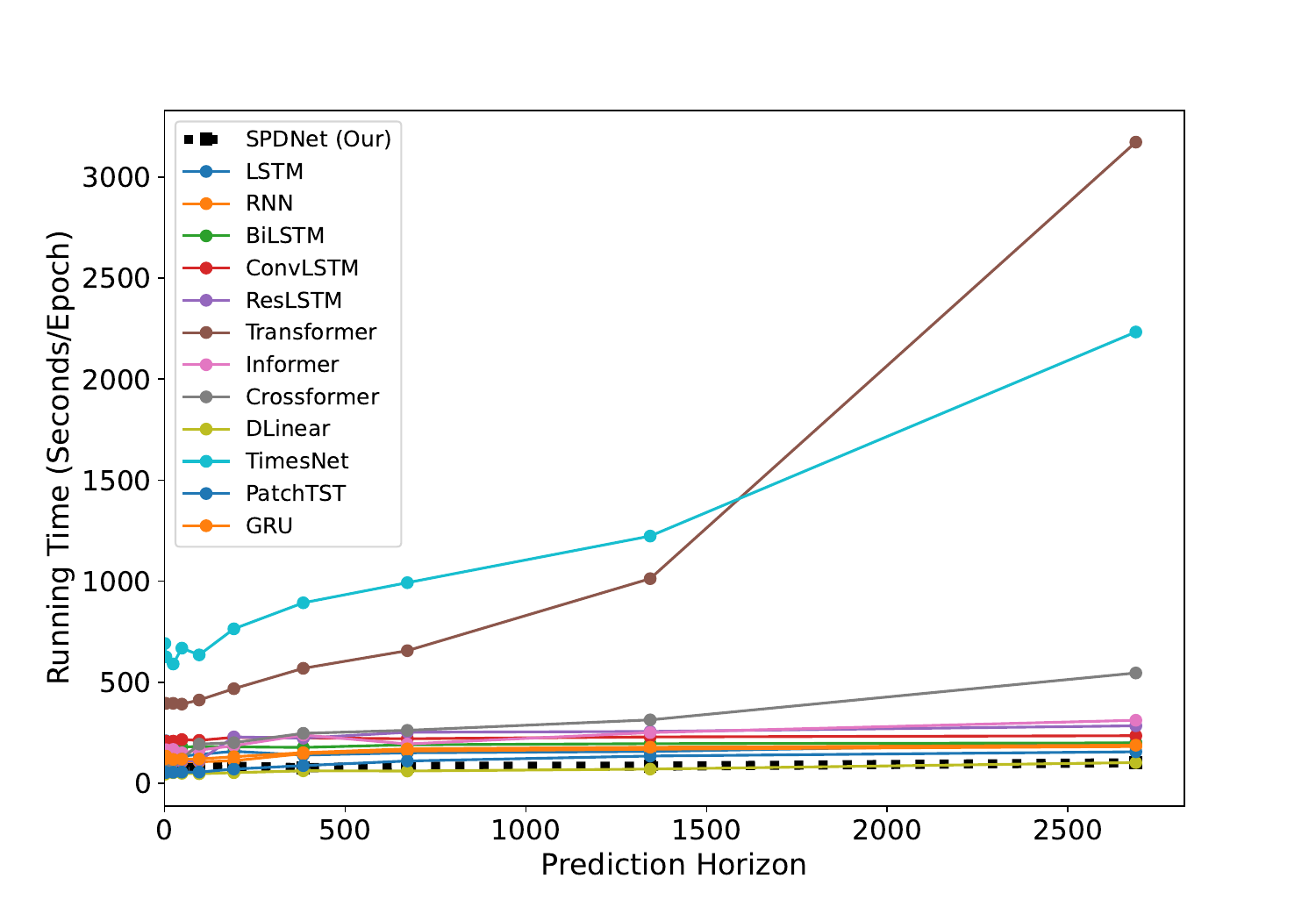} 
\caption{Average training time (seconds) per epoch for different models. All experiments are trained with an input length of 720, batch size of 32, and prediction horizons of \{1,4,24,48,96,192,384,672,1344,2688\}.}
\label{Figure5}
\end{figure}
\section{Experiments}

\subsection{Dataset and Experiment Setup}
\subsubsection{Datasets} The SPDNet model is evaluated using load data of six houses collected from October 25, 2020, to March 10, 2024 from a city in Kansas, USA. Each dataset contains 118,356 rows with 15-minute intervals, and includes weather variables such as air temperature, relative humidity, wind speed, and wind direction to provide context for the analysis. The data was split into 70\% for training, 10\% for validation, and 20\% for testing. To standardize the input features, normalization was performed using the StandardScaler, which standardizes each feature by removing the mean and scaling to unit variance as
\begin{equation}
    z = \frac{x - \mu}{\sigma}
\end{equation}
where $z$ is the normalized value, $x$ is the original value, $\mu$ is the mean, and $\sigma$ is the standard deviation of a feature. Importantly, the scaler was applied solely to the training data to prevent information leakage from the validation and test sets. 

\subsubsection{Baselines} The performance of SPDNet is evaluated against 12 baseline models to ensure a comprehensive analysis across a variety of forecasting methods. This evaluation includes traditional models such as RNN, LSTM, GRU, BiLSTM, ConvLSTM, ResLSTM, and Transformer, as well as advanced techniques like Informer, Crossformer, DLinear, PatchTST, and TimesNet.
 
\subsubsection{Experiment Settings} To ensure a fair comparison, all models were uniformly evaluated on input lengths of \(S=\{96, 380, 720\}\) and prediction horizons of \(P=\{1, 4, 24, 48, 96\}\), with a fixed batch size of 32. A thorough hyperparameter optimization was performed using residential load 1, with the best settings subsequently applied to all other datasets. In addition, for performance evaluation, Mean Squared Error (MSE) and Mean Absolute Error (MAE) metrics were used.

\subsubsection{Implementation Details} All experiments were conducted using PyTorch on a system equipped with an NVIDIA RTX A4000 GPU and an Intel Xeon E5-2680 v3 CPU. The environment also included 16 GB of GPU memory and a multi-core CPU setup to handle extensive computational demands efficiently.

\subsection{Results}
The forecasting results for residential load 1 are provided in Table \ref{tab:mse_comparison}. Additionally, Figure \ref{Figure4} depicts the predicted and actual electricity demand by all models for load 1 and \(S = 96\) and \(P = 96\). The results demonstrate that SPDNet consistently achieves lower MSE and MAE across various prediction horizons and sequence lengths compared to traditional and advanced deep learning models. Notably, at shorter prediction horizons \{1, 4, 24\}, SPDNet shows significant improvements in forecasting accuracy, underscoring its efficiency in capturing short-term fluctuations. As the sequence length increases, SPDNet's performance excels, showcasing its robust ability to analyze extensive historical data through its sophisticated temporal decomposition and attention mechanisms. This capability enables superior handling of long-term trends and periodic fluctuations. 

Among the recursive models (RNN, GRU, and LSTM), the GRU has relatively good performance across MSE and MAE, suggesting it strikes a good balance between complexity and performance. Notably, RNN outperforms LSTM in most of the experiments. The performance of LSTM is significantly improved by making it bidirectional (BiLSTM), integrating convolutional layers (ConvLSTM), and incorporating residual connections (ResLSTM). BiLSTM enhances LSTM by processing data in both forward and backward directions, allowing it to capture dependencies from both past and future contexts. ConvLSTM introduces convolutional layers within the LSTM architecture, enabling it to capture spatial features in time series data, which is particularly beneficial for tasks where the input data includes spatial dimensions. ResLSTM integrates residual connections that help in combating the vanishing gradient problem by allowing gradients to flow through the network more effectively, leading to better performance. 

PatchTST, TimesNet, and DLinear demonstrate strong performance among advanced time series forecasting models. PatchTST leads due to its innovative use of patch-based input processing combined with a Transformer structure, effectively capturing both local and global dependencies. TimesNet follows closely, benefiting from its hierarchical periodical architecture that processes data through multiple 2D convolutional layers to enhance temporal and spatial resolutions. Lastly, DLinear, by utilizing a linear attention mechanism, achieves better performance compared to other advanced forecasting models like Transformer, Informer, and Crossformer.

Figure \ref{Figure5} illustrates training times per epoch for all models with a sequence length of 720 across various prediction horizons \{1, 4, 24, 48, 96, 192, 384, 672, 1344, 2688\}. Compared to competitive models such as PatchTST, DLinear, and TimesNet, SPDNet demonstrates significantly lower average training times per epoch. Notably, as the prediction horizon increases, SPDNet maintains consistent training times, demonstrating its efficiency and stability across different forecasting scales. This performance underscores SPDNet's advanced computational architecture, which is pivotal for enhancing the efficiency and scalability of forecasting models.

Performance of SPDNet is further validated across five additional residential loads to confirm its robustness and generalizability. Using a sequence length of 96 and prediction horizons of \{1, 4, 24\}, SPDNet was trained with the identical parameters and hyperparameters utilized for Load 1. The results are presented in Table \ref{tab:performance_metrics}. SPDNet consistently demonstrates superior performance across various prediction horizons and loads, significantly outperforming LSTM, one of the most powerful models for time series forecasting. 

The extensive evaluations across multiple residential loads and various forecasting models have consistently highlighted SPDNet's superior performance in terms of accuracy and computational efficiency. SPDNet's advanced architectural features enable it to capture intricate temporal variation within residential electrcity demkand. The robustness and scalability demonstrated by SPDNet suggest its potential as a highly effective tool for real-time residential load forecasting.

\begin{table*}[htbp]
\centering
\normalsize 
\renewcommand{\arraystretch}{1.0} 

\caption{Forecasting results for load 1 with normalized values, using a sequence length of $S = {720, 384, 720}$ and prediction horizons $P = {1, 4, 24, 48, 98}$. \textbf{Bold} values indicate the best performance, and \underline{underlined} values denote the second-best.}
\resizebox{\textwidth}{!}{ 
    \begin{tabular}{l|l|ccccc|ccccc|ccccc}
\hline
Model & Metric & \multicolumn{5}{c|}{S=96} & \multicolumn{5}{c|}{S=384} & \multicolumn{5}{c}{S=720} \\ 
\cline{3-17}
       &        & \multicolumn{5}{c|}{Prediction Horizon} & \multicolumn{5}{c|}{Prediction Horizon} & \multicolumn{5}{c}{Prediction Horizon} \\
\cline{3-17}
       &        & 1 & 4 & 24 & 48 & 96 & 1 & 4 & 24 & 48 & 96 & 1 & 4 & 24 & 48 & 96 \\
\hline
SPDNet(Ours) & MSE & \textbf{0.010} & \textbf{0.024} & \underline{0.052} & \textbf{0.056} & 0.064      & \textbf{0.010} & \textbf{0.023} & \textbf{0.050} & \textbf{0.055} & 0.062            & \textbf{0.010} & \textbf{0.024} & \textbf{0.048} & \textbf{0.054} & \textbf{0.057}  \\
              & MAE & \textbf{0.050} & \underline{0.075} & 0.112 & \underline{0.118} & 0.126            & \textbf{0.052} & \textbf{0.077} & \underline{0.118} & \underline{0.120} & 0.148      & \textbf{0.052} & \underline{0.077} & 0.115 & \underline{0.122} & 0.130  \\
\hline
RNN         & MSE & 0.054 & 0.189 & 0.934 & 0.621 & 0.486        & 0.052 & 0.181 & 0.935 & 0.949 & 0.496      & 0.051 & 0.181 & 0.938 & 0.945 & 0.485 \\
            & MAE & 0.217 & 0.343 & 0.945 & 0.685 & 0.482        & 0.214 & 0.334 & 0.945 & 0.952 & 0.472      & 0.210 & 0.346 & 0.947 & 0.950 & 0.469 \\
            
\hline
LSTM        & MSE & 0.029 & 0.26 & 0.970 & 0.990 & 0.456        & 0.017 & 0.254 & 0.966  & 0.995 & 0.556      & 0.031 & 0.251 & 0.969 & 1.000 & 0.568  \\
            & MAE & 0.142 & 0.35 & 0.962 & 0.972 & 0.511        & 0.100 & 0.353 & 0.960  & 0.974 & 0.525      & 0.148 & 0.355 & 0.962 & 0.977 & 0.534 \\
\hline
GRU         & MSE & 0.017 & 0.161 & 0.843 & 0.398 & 0.364        & 0.015 & 0.177 & 0.844 & 0.381 & 0.477      & 0.016 & 0.160 & 0.856 & 0.320 & 0.448  \\
            & MAE & 0.102 & 0.330 & 0.883 & 0.558 & 0.494        & 0.095 & 0.302 & 0.883 & 0.553 & 0.550      & 0.096 & 0.291 & 0.889 & 0.463 & 0.535 \\
\hline
BiLSTM      & MSE & 0.020 & 0.087 & 0.830  & 0.803 & 0.475       & 0.020 & 0.084 & 0.821 & 0.870 & 0.472      & 0.020 & 0.084 & 0.817 & 0.731 & 0.423  \\
            & MAE & 0.108 & 0.215 & 0.889  & 0.726 & 0.459       & 0.110 & 0.216 & 0.884 & 0.810 & 0.454      & 0.106 & 0.208 & 0.882 & 0.678 & 0.429 \\
\hline
ConvLSTM    & MSE & 0.023 & 0.165 & 1.000 & 0.790 & 0.462        & 0.022 & 0.150 & 1.000 & 1.065 & 0.395      & 0.022 & 0.147 & 1.013 & 0.863 & 0.439 \\
            & MAE & 0.125 & 0.271 & 0.980 & 0.706 & 0.452        & 0.118 & 0.265 & 0.981 & 1.010 & 0.426      & 0.118 & 0.275 & 0.984 & 0.717 & 0.432  \\
\hline
ResLSTM     & MSE & 0.020 & 0.198 & 0.904 & 0.917 & 0.899        & 0.024 & 0.188 & 1.308 & 1.305 & 0.982      & 0.200 & 0.897 & 0.920 & 0.899 & 0.899 \\
            & MAE & 0.077 & 0.394 & 0.926 & 0.933 & 0.925        & 0.108 & 0.373 & 1.099 & 1.098 & 0.892      & 0.081 & 0.392 & 0.922 & 0.935 & 0.925\\
\hline           
Transformer  & MSE & 0.026 & 0.184 & 0.588 & 0.583 & 0.535       & 0.025 & 0.160 & 0.592 & 0.642 & 0.538      & 0.124 & 0.329 & 0.574 & 0.801 & 0.738 \\
             & MAE & 0.108 & 0.255 & 0.737 & 0.734 & 0.700       & 0.100 & 0.342 & 0.733 & 0.764 & 0.702      & 0.579 & 1.265 & 2.055 & 2.853 & 2.618 \\
\hline           
Informer     & MSE & 0.036 & 0.062 & 0.563 & 0.762 & 0.763       & 0.027 & 0.074 & 0.440 & 0.730 & 0.639      & NaN & 0.194 & 0.524 & 0.335 & 0.819 \\
             & MAE & 0.122 & 0.185 & 0.639 & 0.769 & 0.760       & 0.092 & 0.205 & 0.604 & 0.754 & 0.658      & NaN & 0.403 & 0.652 & 0.523 & 0.819 \\

\hline           
Crossformer  & MSE & \underline{0.012} & 0.031 & 0.146 & 0.149 & 0.168       & \underline{0.011} & 0.025 & 0.084 & 0.113 & 0.133      & \underline{0.011} & 0.029 & 0.061 & NAN & 0.116 \\
             & MAE & 0.067 & 0.114 & 0.348 & 0.346 & 0.375                   & 0.055 & 0.087 & 0.234 & 0.284 & 0.328      & 0.060 & 0.109 & 0.175 & NAN & 0.239 \\

\hline           
Dlinear      & MSE & \underline{0.012} & \underline{0.025} & 0.062 & 0.070 & 0.078       & 0.012 & \underline{0.024} & \textbf{0.050} & \underline{0.056} & 0.063      & 0.012 & \textbf{0.024} & 0.052 & \underline{0.056} & \underline{0.058} \\
             & MAE & 0.054 & 0.076 & 0.172 & 0.186 & 0.207                               & \underline{0.053} & \textbf{0.077} & 0.119 & 0.131 & 0.162                  & \underline{0.054} & \textbf{0.071} & 0.128 & 0.137 & 0.137 \\

\hline
PatchTST     & MSE & 0.020 & 0.031 & \underline{0.052} & \underline{0.058} & \textbf{0.059}       & 0.023 & 0.034 & 0.053 & \underline{0.058} & \underline{0.060}         & 0.025 & 0.034 & 0.052 & \underline{0.056} & 0.060 \\
             & MAE & 0.068 & 0.080 & \underline{0.105} & \textbf{0.110} & \textbf{0.112}          & 0.076 & 0.091 & \textbf{0.111} & \textbf{0.116} & \textbf{0.119}      & 0.083 & 0.093 & 0.109 & \textbf{0.118} & \textbf{0.120} \\

\hline
TimesNet     & MSE & \underline{0.012} & 0.027 & \textbf{0.051} & \underline{0.058} & \underline{0.060}                & 0.014 & 0.030 & \underline{0.052} & \underline{0.056} & \textbf{0.058}      & 0.016 & \underline{0.030} & 0.054 & 0.064 & \underline{0.058} \\
             & MAE & \underline{0.053} & \textbf{0.074} & \textbf{0.104} & 0.114 & \underline{0.116}                   & 0.059 & \underline{0.081} & \textbf{0.111} & \underline{0.120} & \textbf{0.123}      & 0.063 & 0.085 & 0.118 & 0.131 & \underline{0.128} \\




\hline
\hline
\end{tabular}
} 

\label{tab:mse_comparison}
\end{table*}

\section{Conclusion}
This paper addressed the substantial challenges associated with modeling the complex temporal variation of residential electricity demand, characterized by seasonality, periodicity, long-term dependencies, and sharp fluctuations. To tackle these challenges, the paper introduced the Seasonal-Periodic Decomposition Network (SPDNet), a novel deep learning framework, for enhanced forecasting accuracy. SPDNet is composed of two principal modules: the Seasonal-Trend Decomposition Module (STDM) and the Periodical Decomposition Module (PDM). The STDM effectively decomposes the input data into trend, seasonal, and residual components, enabling a detailed analysis of underlying temporal patterns. Simultaneously, the PDM utilizes the Fast Fourier Transform to identify dominant periods, transforming 1D input data into structured 2D tensors. These tensors are then processed through three parallel submodules: a 1D convolution for detecting rapid fluctuations, a transformer-based encoder for capturing global patterns, and a 2D convolution to model interactions between periods. Extensive simulation results demonstrate SPDNet’s superior performance in both efficiency and accuracy compared to traditional and advanced time series forecasting models. Future work could extend SPDNet to enhance the detection of rare events, focusing on scenarios where anomalous or infrequent patterns significantly impact forecasting accuracy. Furthermore, a deeper examination of the model’s strengths and limitations beyond aggregate metrics like MSE and MAE could provide more nuanced insights into its performance.
\begin{table}[!t]
\caption{Performance Metrics (MSE and MAE) for SPDNet and LSTM Models Across Various Prediction Horizons for Different Loads.\label{tab:performance_metrics}}
\centering
\footnotesize
\renewcommand{\arraystretch}{1.0} 
\begin{tabular}{|>{\centering\arraybackslash}m{0.9cm}|>{\centering\arraybackslash}m{0.6cm}|ccc|ccc|}
\hline
Load & Metric & \multicolumn{3}{c|}{SPDNet} & \multicolumn{3}{c|}{LSTM} \\ \cline{3-8}
     &        & P=1 & P=4 & P=24 & P=1 & P=4 & P=24 \\ \hline
Load 2 & MSE   & 0.001 & 0.002 & 0.005 & 0.006 & 0.038 & 0.302 \\ 
       & MAE   & 0.022 & 0.028 & 0.053 & 0.053 & 0.155 & 0.440 \\ \hline
Load 3 & MSE   & 0.002 & 0.004 & 0.011 & 0.008 & 0.011 & 0.024 \\ 
       & MAE   & 0.019 & 0.025 & 0.048 & 0.048 & 0.062 & 0.100 \\ \hline
Load 4 & MSE   & 0.039 & 0.102 & 0.260 & 0.055 & 0.240 & 0.732 \\ 
       & MAE   & 0.153 & 0.236 & 0.379 & 0.179 & 0.370 & 0.708 \\ \hline
Load 5 & MSE   & 0.010 & 0.026 & 0.072 & 0.019 & 0.084 & 0.219 \\ 
       & MAE   & 0.060 & 0.082 & 0.150 & 0.103 & 0.222 & 0.368 \\ \hline
Load 6 & MSE   & 0.015 & 0.042 & 0.124 & 0.022 & 0.092 & 0.362 \\ 
       & MAE   & 0.083 & 0.140 & 0.256 & 0.107 & 0.223 & 0.534 \\ \hline
\end{tabular}
\end{table}
\section*{Acknowledgments}
This research was conducted with funding from NSF through award No. 2225341.
\bibliographystyle{IEEEtran}
\bibliography{ref}

\end{document}